\pdfoutput=1

\documentclass[11pt]{article}

\usepackage{acl}

\usepackage{times}
\usepackage{latexsym}

\usepackage[T1]{fontenc}

\usepackage[utf8]{inputenc}

\usepackage{microtype}

\usepackage{inconsolata}

\usepackage{bm}
\usepackage{tikz}
\usepackage{array}
\usepackage{float}
\usepackage{xcolor}
\usepackage{bbding}
\usepackage{xspace}
\usepackage{amssymb}
\usepackage{amsmath}
\usepackage{tabularx}
\usepackage{enumitem}
\usepackage{graphicx}
\usepackage{multirow}
\usepackage{multicol}
\usepackage{makecell}
\usepackage{listings}
\usepackage{setspace}
\usepackage{booktabs}
\usepackage{hyperref}
\usepackage{subfigure}

\usepackage[most]{tcolorbox}
\usepackage[ruled]{algorithm2e}

\setlist[itemize]{leftmargin=5mm, itemsep=0mm}

\newcommand{\eg}{\emph{e.g.,}\xspace}

\newcommand{\name}{SpreadsheetAgent\xspace}
\newcommand{\first}{Structure Extraction Stage\xspace}
\newcommand{\second}{Solving Stage\xspace}

\definecolor{title}{rgb}{0.2, 0.2, 0.2}
\definecolor{background}{rgb}{1, 0.975, 0.975}

\definecolor{darkgreen}{RGB}{0,100,0}

\title{Towards Robust Real-World Spreadsheet Understanding with \\ Multi-Agent Multi-Format Reasoning}

\author{
  Houxing Ren$^1$ \quad Mingjie Zhan$^2\footnotemark[1]$ \quad Zimu Lu$^1$ \quad Ke Wang$^1$ \quad Yunqiao Yang$^1$  \\
  \textbf{Haotian Hou}$^2$ \quad \textbf{Hongsheng Li}$^{1,3,4}$\thanks{Corresponding author.} \\
  $^1$CUHK MMLab ~ $^2$SenseTime Research ~ $^3$Shenzhen Loop Area Institute ~ $^4$CPII under InnoHK ~ \\
  renhouxing@gmail.com \quad zhanmingjie@sensetime.com \quad hsli@ee.cuhk.edu.hk
}

\begin{document}

\maketitle

\begin{abstract}
Spreadsheets are central to real-world applications such as enterprise reporting, auditing, and scientific data management. Despite their ubiquity, existing large language model based approaches typically treat tables as plain text, overlooking critical layout cues and visual semantics. Moreover, real-world spreadsheets are often massive in scale, exceeding the input length that LLMs can efficiently process. To address these challenges, we propose \name, a two-stage multi-agent framework for spreadsheet understanding that adopts a step-by-step reading and reasoning paradigm. Instead of loading the entire spreadsheet at once, \name incrementally interprets localized regions through multiple modalities, including code execution results, images, and \LaTeX{} tables. The method first constructs a structural sketch and row/column summaries, and then performs task-driven reasoning over this intermediate representation in the \second. To further enhance reliability, we design a verification module that validates extracted structures via targeted inspections, reducing error propagation and ensuring trustworthy inputs for downstream reasoning. Extensive experiments on two spreadsheet datasets demonstrate the effectiveness of our approach. With GPT-OSS-120B, \name achieves 38.16\% on Spreadsheet Bench, outperforming the ChatGPT Agent baseline (35.27\%) by 2.89 absolute points. These results highlight the potential of \name to advance robust and scalable spreadsheet understanding in real-world applications. Code is available at \url{https://github.com/renhouxing/SpreadsheetAgent}.
\end{abstract}

\section{Introduction} \label{sec:intro}

Spreadsheets are among the most widely used data formats in real-world applications, ranging from enterprise reporting and financial auditing to scientific data management and governmental records. Their structured nature makes them essential for storing, organizing, and analyzing large volumes of information. With the rapid development of large language models~(LLMs)~\cite{GPT42023ABS230308774, LLama22023ABS230709288, Mixtral2023ABS240104088, Qwen252024Yang}, table processing has emerged as an active research direction. A series of new systems, such as TableGPT~\cite{TableGPT2024Peng, TableGPT22025AoFeng}, Chain-of-Table~\cite{ChainOfTable2024Zilong}, TaPERA~\cite{TaPERA2024Yilun}, SheetAgent~\cite{SheetAgent2025Yibin}, and SheetMind~\cite{SheetMind2025Ruiyan}, have been proposed to improve tasks like table understanding, reasoning, and interaction. These models highlight the potential of LLMs to enhance efficiency in practical scenarios, including automated reporting, statistical analysis, and intuitive spreadsheet manipulation.

However, despite these advances, most existing approaches represent tables as pure text, \eg Markdown~\cite{TaPERA2024Yilun, TableGPT2024Peng}, HTML~\cite{Structure2024Yuan, RealHiT2025Pengzuo}, or \LaTeX{}~\cite{Structure2024Yuan, RealHiT2025Pengzuo}, with only a few exploring image-based inputs~\cite{MMTable2024Mingyu, SynTab2025Bangbang}. Real-world spreadsheets, by contrast, are far more complex: they contain hierarchical headers, multiple sheets, and rich visual cues such as font colors, shaded cells, and borders. These stylistic elements carry semantic information that text-only formats cannot fully capture. Moreover, practical spreadsheets often span thousands of rows and columns, exceeding the context length that current LLMs can efficiently handle. As a result, existing methods struggle to represent both the structural complexity and the scale of spreadsheets, limiting their effectiveness in downstream tasks such as question answering and knowledge extraction.

To overcome these challenges, we propose \name, a two-stage multi-agent framework built on step-by-step reading and reasoning, where an extraction agent collaborates with a vision range agent and a \LaTeX{} range agent to process spreadsheets incrementally. Rather than ingesting the entire spreadsheet in one pass, our framework allows agents to iteratively feed the model with small regions in multiple formats, \eg code execution results, images, and \LaTeX{} tables. In this way, the model reads and reasons incrementally under a tight context budget. The two stages operationalize this paradigm: the \first constructs a structural sketch and row/column summaries through iterative region inputs; the \second performs task-driven reasoning by consulting and, when necessary, extending this sketch. Compared with previous approaches, \name preserves layout semantics while avoiding the need to load the entire table at once.

More concretely, in the \first, the extraction agent drives the exploration: it scans the spreadsheet, identifies structural cues such as hierarchical headers, merged cells, and multi-sheet organization, and assembles concise row-level and column-level summaries. To support this process, the extraction agent issues specialized requests to two assistants who answer queries and verify representations while delegating conversions to auxiliary tools. The vision range agent formulates visual queries over selected ranges, invokes the image-conversion module to render snapshots when needed, interprets visual cues, and performs vision-based verification by cross-checking the evolving sketch against the rendered views. The \LaTeX{} range agent formulates structural queries over ranges, invokes the \LaTeX{}-conversion module to obtain symbolic tables that preserve hierarchical headers and alignment, interprets the resulting structures, and conducts \LaTeX{}-based verification for faithfulness. Through this division of labor, the agents jointly construct a compact intermediate representation that faithfully captures both spreadsheet content and layout, providing a reliable foundation for downstream reasoning.

Our main contributions are as follows:
\begin{itemize}[leftmargin=5mm, itemsep=0mm]
    \item We propose \name, a novel two-stage multi-agent framework for spreadsheet understanding, which adopts a step-by-step reading and reasoning paradigm, enabling progressive interpretation without exceeding the context-length limitations of LLMs.  
    \item We design a structural exploration and reasoning pipeline that processes spreadsheets through localized region-based inputs in multiple formats, including code execution results, image snippets, and \LaTeX{} expressions, thereby capturing a wide range of semantics.  
    \item Extensive experiments demonstrate the effectiveness of \name. On SpreadsheetBench, our method achieves 38.16\% with GPT-OSS-120B, substantially outperforming the ChatGPT Agent\footnote{\url{https://openai.com/index/introducing-chatgpt-agent/}} (35.27\%). This 2.89 absolute improvement highlights the capability of \name in handling complex spreadsheet reasoning tasks.
\end{itemize}
\section{Related Work} \label{sec:rel}

\subsection{Table Understanding}

A large body of research has sought to improve how language models understand and reason over tables, often by designing specialized representations or reasoning procedures. 
Chain-of-Table~\cite{ChainOfTable2024Zilong} introduced step-by-step tabular reasoning by iteratively generating operations that transform tables into a chain of intermediate representations, enabling systematic decomposition of complex queries. 
SheetMind~\cite{SheetMind2025Ruiyan} proposed an LLM-powered multi-agent framework for spreadsheet automation via natural language, which uses a Manager Agent, an Action Agent, and a Reflection Agent integrated with Google Sheets.
Similarly, SheetAgent~\cite{SheetAgent2025Yibin} proposed an LLM-based agent with Planner, Informer, and Retriever modules for enhanced spreadsheet reasoning and manipulation.
ReAcTable~\cite{ReAcTable2025Yunjia} extended this idea with external tool augmentation and a voting mechanism, improving robustness in table question answering. 
TaPERA~\cite{TaPERA2024Yilun} tackled compositional queries by decomposing them into sub-questions, executing Python programs to retrieve facts from tables, and synthesizing long-form answers from program outputs. 
Beyond symbolic reasoning, multimodal methods have also emerged. TabPedia~\cite{TabPedia2024Weichao} unified multiple visual table understanding tasks via concept synergy, leveraging dual encoders for high- and low-resolution image features. Similarly, RealHiT~\cite{RealHiT2025Pengzuo} focused on hierarchical headers, prompting the model to first generate a header tree before performing reasoning. 
More recently, general frameworks have been proposed to address diverse challenges in table reasoning. TableMaster~\cite{TableMaster2025Lang} introduced a unified system incorporating table-of-focus, verbalization, program-aided reasoning, and adaptive strategies to handle different table tasks. ROT~\cite{RoT2025Xuanliang} proposed a row-by-row traversal strategy with post-traversal reflection, aiming to reduce hallucinations while maintaining efficiency. 

\subsection{Fine-tuning for Table Understanding}

Another major research direction is to adapt pretrained models to table tasks via fine-tuning or instruction tuning. TableGPT~\cite{TableGPT2024Peng} pioneered this by synthesizing diverse table-related tasks from real-world tables and applying instruction tuning, yielding improved generalization. 
This inspired follow-up work such as TableGPT2~\cite{TableGPT22025AoFeng}, Table-LLaVA~\cite{MMTable2024Mingyu}, and SynTab~\cite{SynTab2025Bangbang}, which extended the focus to multimodal table understanding through the construction of large-scale instruction-tuning datasets.
Motivated by the strong performance of DeepSeek-R1~\cite{deepseekr12025deepseek}, reinforcement learning (RL) has also become a central paradigm for enhancing reasoning in language models, leading to a surge of RL-based approaches. In the table domain, two representative lines of work have recently been proposed. 
\citet{TableR12025Rihui} focuses on small language models, introducing layout-aware pretraining and a mix-paradigm GRPO variant to improve robustness and consistency, achieving substantial gains over base models. 
In contrast, \citet{TableR12025Zhenhe} combined region-guided reasoning with table-aware GRPO to enhance efficiency and accuracy, even surpassing much larger baselines. These studies highlight the promise of RL in advancing language-based tabular reasoning.

\section{Methodology} \label{sec:met}

\begin{figure*}[t]
    \centering
    \includegraphics[width=\textwidth]{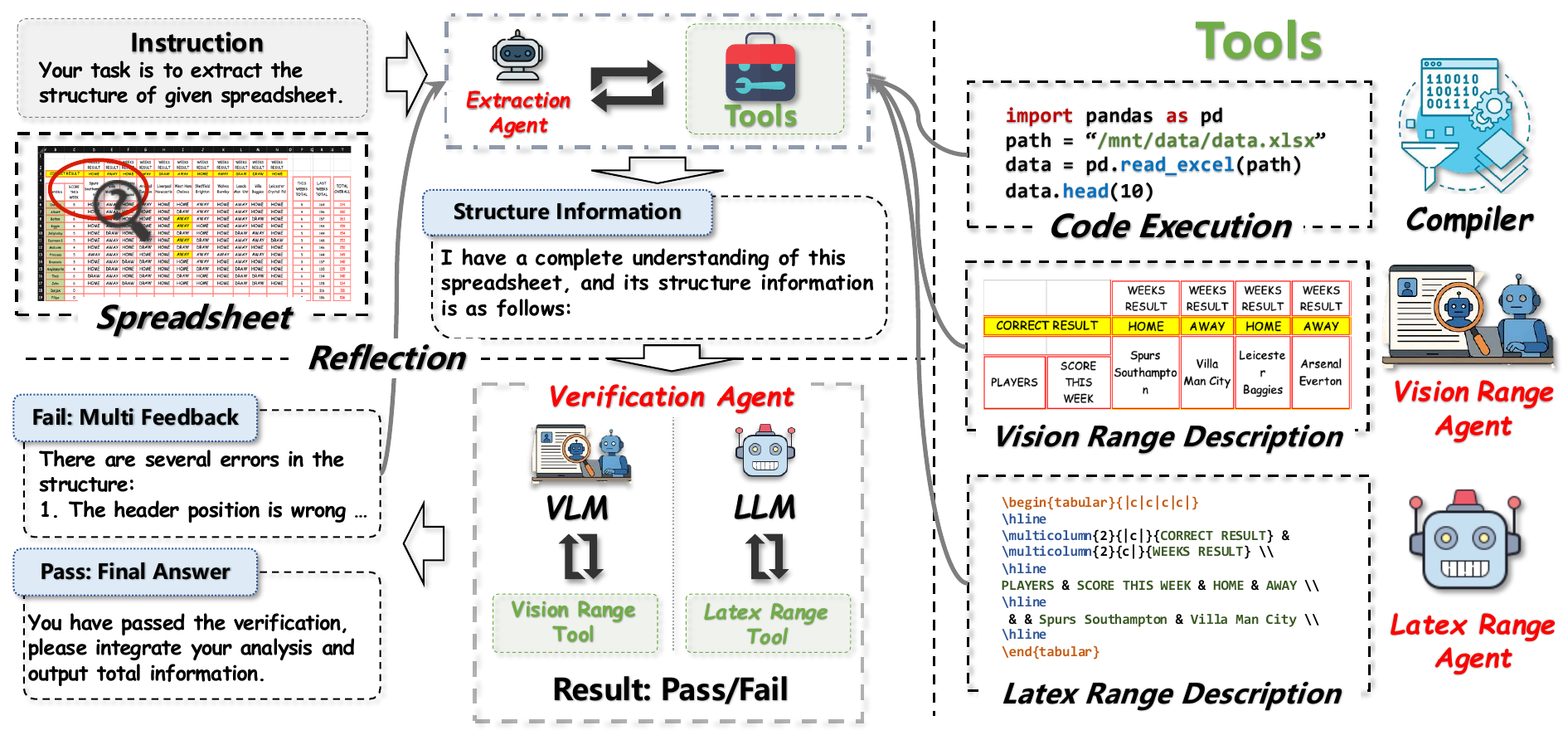}
    \caption{Overview of the proposed multi-agent framework for spreadsheet understanding. An Extraction Agent incrementally parses the spreadsheet with the help of code execution, vision range description, and \LaTeX{}  range description tools, while a Verification Agent cross-checks results using VLM-based and LLM-based range descriptions. Through iterative reflection and feedback, the system produces a faithful structural representation that preserves layout semantics under tight context budgets.}
    \label{fig:overview}
\end{figure*}

In this section, we present our framework for spreadsheet understanding, shown in Figure~\ref{fig:overview}. We begin by describing how to extract structural and semantic information from spreadsheets through a dedicated extraction module. Next, we present how to verify the extracted results against the original table using a verification module to enhance reliability. Finally, we provide an overview of the entire pipeline.

\subsection{Extraction Module}

The extraction module targets the full diversity of real-world spreadsheets rather than treating tables as flat matrices. In practice, spreadsheets frequently include (1) hierarchical headers that encode layered categorical semantics, (2) complex row/column semantics such as group totals, stubs, and sectioned indices. These layout signals determine how cell values should be interpreted and aggregated; ignoring them leads to misaligned schemas and brittle reasoning. Therefore, our design instructs the model to explicitly detect and preserve these structures during the extraction process.

Specifically, we adopt a prompt-driven pipeline that guides the LLM to identify structural cues and to summarize them alongside localized content. Concretely, the prompt enumerates the expected phenomena (hierarchical headers, merged cells, sheet-level sectioning, formatting-based semantics such as font color or shading) and asks the model to return a normalized schema that binds layout to semantics. This step converts loosely formatted spreadsheets into a faithful intermediate description while remaining robust to large tables via region-wise processing.
To further ensure reliability for subsequent modules, the model is required to emit outputs in a YAML format. We choose this format because it is human-readable, captures nested structure with minimal loss, and is straightforward to parse programmatically. We empirically observe that the output format materially affects downstream performance: structured YAML reduces ambiguity, stabilizes parsing, and improves compatibility with task-specific reasoners compared to free-form text.

In addition to structured YAML outputs, which improve downstream interpretability and stability, we further equip the system with three auxiliary tools that enhance the model’s reasoning over challenging spreadsheet phenomena. These tools complement the schema-driven approach by providing precise access to raw values, visual cues, and structural signals that are otherwise difficult to capture through text prompting alone.
\begin{itemize}[leftmargin=5mm, itemsep=0mm]
  \item \textbf{Code Execution.} The system can directly execute Python code written by the model, enabling precise parsing of raw values, numerical computation, or programmatic validation. Unlike conversion-based tools, this mechanism provides exact results for arithmetic operations and custom logic that go beyond approximate text reasoning.
  \item \textbf{Vision Range Description Agent.} Given a selected range, the system converts it into an image, and then a vision-language model is prompted with the image and a question to extract semantic information from visual cues such as colors, borders.
  \item \textbf{\LaTeX{} Range Description Agent.} Given a selected range, the system converts it into a \LaTeX{} table through code, and then a language model is prompted with the \LaTeX{} representation and a question to recover structural information such as hierarchical headers or merged cells.
\end{itemize}
By combining structure-aware prompting, a YAML-based intermediate, and tool-assisted mechanisms (code execution, vision, and \LaTeX{}), the extraction module produces a compact representation that retains both content fidelity and layout semantics. This representation serves as the foundation for subsequent verification and task-driven reasoning.

\subsection{Verification Module}

For many tasks, verification is often easier than direct problem solving. Instead of asking the model to generate a complete interpretation from scratch, it is typically more reliable to let the model check whether an existing extraction is correct. This motivates our design of a verification module, which improves accuracy by validating the outputs of the extraction module before they are consumed in downstream reasoning.

Specifically, given the entire spreadsheet and its extracted YAML representation, the verification agent checks whether the extraction faithfully reflects the original table. Instead of reprocessing everything at once, which would exceed context limits for large spreadsheets, the agent selectively focuses on uncertain or structurally complex regions. It actively decides which regions to convert into alternative formats (image or \LaTeX{}) for further inspection, thereby reducing error propagation and ensuring more reliable inputs for downstream reasoning.

\begin{itemize}[leftmargin=5mm, itemsep=0mm]
  \item \textbf{Vision Verification Agent.} The agent may convert a selected region into an image and query a vision-language model to check whether the extracted schema matches the visual layout. Detected issues—such as missing merged cells, misinterpreted colors, or overlooked chart elements—are reported with suggested corrections.
  \item \textbf{\LaTeX{} Verification Agent.} The agent may alternatively render a region as a \LaTeX{} table and prompt a language model to verify structural fidelity. Features like hierarchical headers, multicolumn or multirow spans, and alignment are cross-checked against the YAML extraction, with any inconsistencies flagged and accompanied by revision proposals.
\end{itemize}
In this design, the verification module does not attempt to reprocess the entire table at once. Instead, it strategically combines the extracted YAML with targeted tool-assisted inspections. By leveraging adaptive region selection and focusing on checking rather than solving, the framework increases robustness and ensures that downstream reasoning is grounded in faithful representations of spreadsheet content and structure.

\begin{algorithm*}[t]
\caption{Incremental Extraction-Verification Loop}
\label{alg:method}
\KwIn{Spreadsheet $S$}
\KwOut{Verified structural representation $Y^\star$}

\BlankLine
$\mathcal{C} \leftarrow \{S,\ \text{Prompt},\ \text{Tool interfaces}\}, Y \leftarrow \emptyset, Y^\star \leftarrow \emptyset$\;

\Repeat{verification succeeds or max-iterations reached}{
    
    \While{the agent invokes a tool}{
        Select tool $T \in \{\texttt{Code}, \texttt{Vision}, \texttt{LaTeX}\}$ and parameters $P$\;
        $O \leftarrow T(S,P)$\;
        Append $O$ to $\mathcal{C}$\;
    }
    $Y \leftarrow \mathcal{C}[-1], \mathit{vision\_pass} \leftarrow \mathrm{false}, \mathit{latex\_pass} \leftarrow \mathrm{false}$\;

    $\mathcal{C}_{v} \leftarrow \{S,\ Y,\ \text{Verify prompt},\ \text{Vision interfaces}\}$\;
    \While{the vision agent invokes a vision tool}{
        Select parameters $P_v$\;
        $R_v \leftarrow \texttt{VisionTool}(S,P_v)$\;
        Append $R_v$ to $\mathcal{C}_{v}$\;
    }
    $\mathit{vision\_pass},\ \Delta_v \leftarrow \texttt{parse\_verification}(\mathcal{C}_{v}[-1])$\;
    
    $\mathcal{C}_{l} \leftarrow \{S,\ Y,\ \text{Verify prompt},\ \text{LaTeX interfaces}\}$\;
    \While{the LaTeX agent invokes a LaTeX tool}{
        Select parameters $P_l$\;
        $R_l \leftarrow \texttt{LaTeXTool}(S,P_l)$\;
        Append $R_l$ to $\mathcal{C}_{l}$\;
    }
    $\mathit{latex\_pass},\ \Delta_l \leftarrow \texttt{parse\_verification}(\mathcal{C}_{l}[-1])$\;
    
    \uIf{$\mathit{vision\_pass}$ and $\mathit{latex\_pass}$}{
        $Y^\star \leftarrow Y$\;
        \textbf{break}\;
    }
    \Else{
        Append $\Delta_v$ and $\Delta_l$ to $\mathcal{C}$\;
    }
}
\Return{$Y^\star$}
\end{algorithm*}

\subsection{Discussion}

As described in Algorithm~\ref{alg:method}, the two modules described above operate in a tightly-coupled, iterative loop rather than as independent components. The extraction module first produces a structured intermediate representation that preserves both spreadsheet content and layout. The verification module then inspects this representation against the original table, identifies potential errors, and proposes corrections. These corrections are fed back into the extraction process, which updates the representation accordingly. The cycle of extraction–verification–refinement continues until the schema achieves sufficient fidelity.

By combining extraction and verification in this iterative manner, our framework establishes a progressive reading-and-checking paradigm. Instead of attempting to process an entire spreadsheet in one step, the system incrementally builds and validates its understanding. This design offers two major benefits: (i) improved scalability, since only localized regions are processed at each step under a limited context budget; and (ii) enhanced accuracy, as verification reduces error propagation and strengthens reliability. Finally, once the representation has been validated, the resulting structured information is directly injected into the context of downstream tasks, ensuring that subsequent reasoning operates on a faithful and semantically enriched view of the spreadsheet.

\section{Experiments} \label{sec:exp}

\begin{table*}[t]
\centering
\caption{Main experimental results on SpreadsheetBench. We report accuracy under both Soft Restriction and Hard Restriction settings, further broken down into cell-level, sheet-level, and overall scores. Numbers for existing methods are taken from the SpreadsheetBench paper.}
\begin{tabular}{l|ccc|ccc}
\toprule
\multirow{2.5}{*}{Model} & \multicolumn{3}{c|}{Soft Restriction ($\uparrow$)} & \multicolumn{3}{c}{Hard Restriction ($\uparrow$)} \\ \cmidrule(lr){2-7}
 & Cell-Level & Sheet-Level & Overall & Cell-Level & Sheet-Level & Overall \\
\midrule
Binder (GPT-3.5) & 1.58 & 0.05 & 1.17 & 0.00 & 0.00 & 0.00 \\
\midrule
Mixtral-8x7B & 3.39 & 4.67 & 3.88 & 2.32 & 3.71 & 2.85 \\
Llama-3-70B & 1.13 & 7.90 & 3.74 & 0.71 & 7.14 & 3.18 \\
\midrule
GPT-3.5 & 3.33 & 13.11 & 7.09 & 2.50 & 9.97 & 5.37 \\
GPT-4o & 13.49 & 22.51 & 16.96 & 10.52 & 17.66 & 13.27 \\
\midrule
SheetCopilot & 16.67 & 10.00 & 14.00 & - & - & - \\
Copilot in Excel & 23.33 & 15.00 & 20.00 & - & - & - \\
ChatGPT Agent & 38.27 & 30.48 & 35.27 & - & - & - \\
\midrule
GPT-OSS-20B & 25.55 & 24.79 & 25.26 & 19.43 & 21.37 & 20.18 \\
\quad w/ TreeThinker & 23.95 & 26.12 & 24.78 & 18.54 & 22.51 & 20.07 \\
\quad w/ \name & \textbf{34.46} & \textbf{28.49} & \textbf{32.16} & \textbf{27.81} & \textbf{25.64} & \textbf{29.25} \\
\midrule
GPT-OSS-120B & 30.78 & 27.64 & 29.57 & 24.96 & 23.93 & 24.56 \\
\quad w/ TreeThinker & 32.14 & 29.82 & 31.25 & 24.42 & 26.50 & 25.22 \\
\quad w/ \name & \textbf{41.30} & \textbf{33.14} & \textbf{38.16} & \textbf{32.80} & \textbf{29.34} & \textbf{31.47} \\
\midrule
Qwen3-30B & 13.61 & 20.89 & 16.41 &  9.45 & 18.23 & 12.83 \\
\quad w/ TreeThinker & 14.38 & 22.22 & 17.40 & 10.16 & 19.09 & 13.60 \\
\quad w/ \name & \textbf{20.62} & \textbf{25.17} & \textbf{22.37} & \textbf{16.04} & \textbf{22.22} & \textbf{18.42} \\
\midrule
Qwen3-235B & 20.50 & 25.83 & 22.55 & 15.33 & 22.51 & 18.09 \\
\quad w/ TreeThinker & 22.99 & 26.50 & 24.34 & 17.29 & 23.65 & 19.74 \\
\quad w/ \name & \textbf{33.10} & \textbf{28.77} & \textbf{31.43} & \textbf{25.31} & \textbf{25.07} & \textbf{25.22} \\
\midrule
Qwen3-Coder-480B & 30.36 & 31.05 & 30.63 & 22.82 & 27.07 & 24.45 \\
\quad w/ TreeThinker & 38.86 & 32.00 & 36.22 & 31.91 & 27.92 & 30.37 \\
\quad w/ \name & \textbf{45.63} & \textbf{35.33} & \textbf{41.67} & \textbf{36.90} & \textbf{31.05} & \textbf{34.65} \\
\midrule
Human Performance & 75.56 & 65.00 & 71.33 & 66.67 & 55.00 & 62.00 \\
\bottomrule
\end{tabular}
\label{tab:spreadsheet}
\end{table*}
\begin{table*}[t]
\centering
\caption{Ablation study of \name using Qwen3-30B as the reasoning model. We examine the contributions of verification, structural sketching, and multi-format tools.}
\begin{tabular}{l|ccc|ccc}
\toprule
\multirow{2.5}{*}{Model} & \multicolumn{3}{c|}{Soft Restriction ($\uparrow$)} & \multicolumn{3}{c}{Hard Restriction ($\uparrow$)} \\ \cmidrule(lr){2-7}
 & Cell-Level & Sheet-Level & Overall & Cell-Level & Sheet-Level & Overall \\
\midrule
\name & 20.62 & \textbf{25.17} & \textbf{22.37} & \textbf{16.04} & \textbf{22.22} & \textbf{18.42} \\
\midrule
\quad w/o Tools \& Verify & 19.61 & 21.08 & 20.18 & 14.44 & 18.52 & 16.01 \\
\qquad w/ JSON & 19.37 & 22.98 & 20.76 & 13.90 & 20.23 & 16.34 \\
\qquad w/o Structure & 18.84 & 21.08 & 19.70 & 14.08 & 17.66 & 15.46 \\
\midrule
\quad w/o Verify & 20.44 & 23.08 & 21.45 & 15.69 & 20.51 & 17.54 \\
\qquad w/o Vision Tool & \textbf{21.03} & 22.13 & 21.45 & 15.33 & 19.37 & 16.89 \\
\qquad w/o Latex Tool & 19.01 & 22.41 & 20.32 & 14.44 & 19.09 & 16.23 \\
\midrule
\quad w/o All & 13.61 & 20.89 & 16.41 & 9.45 & 18.23 & 12.83 \\
\bottomrule
\end{tabular}
\label{tab:ablation}
\end{table*}

In this section, we present extensive experiments to demonstrate the effectiveness of the proposed method and analyze its performance. Due to the limited space, more detailed experiments are presented in Appendix~\ref{sec:app_exp}.

\subsection{Experimental Setup} \label{sec:setup}

\paragraph{Test Dataset.}
We evaluate our method on SpreadsheetBench~\cite{SpreadsheetBench2024Zeyao}, a benchmark constructed from 912 real-world questions collected from Excel user forums. Each question is paired with its original spreadsheet file, which often contains multiple tables, non-standard relational structures, merged cells, and other non-textual elements such as pivot tables, charts, and annotations.
In contrast, many existing benchmarks are constructed with highly regularized tables, such as WikiTQ~\cite{WikiTQ2015Panupong} and TabFact~\cite{TabFact2020Wenhu}, which cannot fully capture the irregular structures and noisy layouts often encountered in practical spreadsheet scenarios. Other benchmarks focus on hierarchical headers, such as HiTab~\cite{HiTab2022Zhoujun} and AIT-QA~\cite{AITQA2022Yannis}, but represent them only in a serialized textual form (\eg JSON), which strips away the layout and formatting cues that are essential for spreadsheet manipulation.

To further evaluate our method under settings that specifically emphasize hierarchical headers, we also conduct experiments on RealHiTBench~\cite{RealHiT2025Pengzuo} in Appendix~\ref {sec:realhit}, which provides Excel files with complex hierarchical headers. This allows us to isolate and analyze the effectiveness of our approach on tasks dominated by header complexity, complementing the broader coverage of SpreadsheetBench.

\paragraph{Implementation Details.} 
In our experiments, we use GLM-4.5V\footnote{\url{https://huggingface.co/zai-org/GLM-4.5V}}~\cite{GLM45V2025Wenyi} as the vision–language model, serving both as the vision conversion tool and the vision verification model, and Qwen3-Coder-480B\footnote{\url{https://huggingface.co/Qwen/Qwen3-Coder-480B-A35B-Instruct}}~\cite{Qwen32025An} as the language model, which acts as the main model, the \LaTeX{} conversion tool, and the \LaTeX{} verification model. For inference acceleration, we adopt vLLM~\cite{vLLM2025Thomas}, using greedy decoding with temperature set to 0 and top-p set to 1. Spreadsheet content is serialized into structured text and appended to the prompt. All models are evaluated with the same prompt template to ensure fairness, using a maximum context length of 4K tokens per round and allowing at most 20 rounds of tool calls in total.
We benchmark all models under identical conditions. For the Spreadsheet Bench, we report accuracy as the evaluation metric, while for the RealHiT Bench, we adopt Exact Match (EM) and F1 scores. All experiments are conducted on 8 NVIDIA H800 GPUs with vLLM, leveraging tensor parallelism and expert parallelism for efficient inference.

\subsection{Evaluation}

\paragraph{Baseline.}
We compare our method against a wide range of baselines, including both general-purpose LLMs (\eg GPT-3.5, GPT-4o, Llama-3) and spreadsheet-specific systems such as SheetCopilot~\cite{SheetCopilot2023Hongxin}, Copilot in Excel\footnote{\url{https://www.microsoft.com/microsoft-copilot}}, and ChatGPT Agent\footnote{\url{https://openai.com/index/introducing-chatgpt-agent/}}. 
For fairness, the results of these methods are directly taken from the original SpreadsheetBench paper. 
In addition, we include TreeThinker~\cite{RealHiT2025Pengzuo} as a strong recent approach for reasoning over hierarchical table structures. 
Human performance is further reported as an upper-bound reference.

\paragraph{Result.}
Table~\ref{tab:spreadsheet} reports the comparison results on SpreadsheetBench, from which we derive several key observations:
(1) All existing systems still fall far short of human performance. Even GPT-4o, one of the most capable proprietary models, only achieves 18.35\% under the soft restriction and 15.02\% under the hard restriction, highlighting the inherent difficulty of real-world spreadsheet reasoning.
(2) Our method establishes new state-of-the-art results. While the ChatGPT Agent was previously the strongest reported system with 35.27\% accuracy under the soft setting, our framework achieves significant improvements. In particular, \name with Qwen3-480B-A35B reaches 41.67\% and with GPT-OSS-120B reaches 38.16\%, surpassing the ChatGPT Agent across different model scales and demonstrating the robustness of our design.
(3) Our framework consistently outperforms TreeThinker. A key limitation of TreeThinker is its exclusive focus on hierarchical headers, whereas our approach incorporates a broader spectrum of spreadsheet cues. By integrating visual signals and \LaTeX{}-based representations, \name preserves richer layout and semantic information, leading to stronger generalization across the diverse structures present in SpreadsheetBench.
(4) Beyond absolute accuracy, the results reveal consistent gains across different model scales. On smaller backbones such as Qwen3-30B, \name improves the overall score by nearly 6 points, while on large-scale models like Qwen3-Coder-480B, the gain exceeds 11 points. This trend confirms that the proposed framework is not restricted to high-capacity models but provides robust benefits across scales.

\subsection{Detailed Analysis}

\subsubsection{Ablation Study}

Here, we check how each component contributes to the final performance. We design several variants to examine the role of different components. Specifically, \underline{w/o Tools \& Verify} disables all external tools and the verification mechanism, and its sub-variants are: \underline{w/ JSON}, which replaces YAML structure with JSON as the table representation; and \underline{w/o Structure}, which does not have any format restrictions. In addition, \underline{w/o Verify} disables the verification process, with two sub-variants: \underline{w/o Vision Tool}, which removes the image conversion module; and \underline{w/o Latex Tool}, which removes the \LaTeX{} conversion module. Finally, \underline{w/o All} eliminates all proposed components, leaving only the base reasoning agent.

Table~\ref{tab:ablation} presents the ablation results of \name on SpreadsheetBench. Based on the results, several important findings can be drawn.
First, the full model achieves the highest overall accuracy (22.37\% / 18.42\%), indicating that the joint use of all modules yields the strongest performance.
Second, verification plays a critical role: removing it (\underline{w/o Verify}) consistently lowers accuracy, while the even larger gap observed in \underline{w/o Tools \& Verify} highlights that tools and verification are mutually reinforcing rather than individually sufficient.
Third, structural sketching proves essential. In the default setting, \underline{w/o Tools \& Verify} produces YAML outputs. We further evaluate \underline{w/ JSON} and \underline{w/o Structure}, which either replace YAML with JSON or discard sketching entirely. The results show that both YAML and JSON formats support relatively strong performance, as explicit structural representations allow downstream models to interpret the data more effectively. In contrast, removing structured outputs severely degrades accuracy, confirming that structural sketching is a key component for spreadsheet understanding.
Fourth, the multi-format tools contribute in complementary ways. Removing the vision tool leads to slight gains in soft cell-level tasks but significantly harms performance under hard settings. Conversely, removing the \LaTeX{} tool consistently reduces accuracy, underscoring its importance for preserving layout and structural fidelity.
Finally, the \underline{w/o All} variant results in a dramatic collapse (16.41\% / 12.83\%), nearly 10 points below the full model, which provides strong evidence that integrating all components into a unified framework is indispensable.

\subsubsection{Effect of Structure Extractor}

To further understand the impact of the structure extractor on the final performance, we conduct an ablation study where the extractor model is replaced while keeping the reasoning model fixed. Specifically, we substitute the Qwen3-480B extractor with two smaller candidates: Qwen3-30B and GPT-OSS-20B. Note that we do not replace GLM-4.5V, because the availability of LLMs that can both handle image inputs and support function calling is still very limited, and there is no comparable smaller-scale alternative available.

\begin{table}[t]
\centering
\caption{Ablation study on the impact of different structure extractors. The multi-modal reasoning model is fixed with GLM-4.5V, while we vary the language reasoning model among Qwen3-480B, Qwen3-30B, and GPT-OSS-20B. }
\setlength{\tabcolsep}{4pt}
\begin{tabular}{l|c|cc}
\toprule
Model & Extractor & Soft & Hard \\
\midrule
Qwen3-30B & Qwen3-480B & \textbf{22.37} & \textbf{18.42} \\
Qwen3-30B & Qwen3-30B & 17.47 & 13.82 \\
Qwen3-30B & GPT-OSS-20B & 19.44 & 15.79 \\
Qwen3-30B & - & 16.41 & 12.83 \\
\midrule
GPT-OSS-20B & Qwen3-480B & \textbf{32.16} & \textbf{29.25} \\
GPT-OSS-20B & Qwen3-30B & 25.71 & 20.83 \\
GPT-OSS-20B & GPT-OSS-20B & 27.85 & 22.59 \\
GPT-OSS-20B & - & 25.26 & 20.18 \\
\bottomrule
\end{tabular}
\label{tab:extractor}
\end{table}

The results in Table~\ref{tab:extractor} clearly show that using an extractor at least as strong as the reasoning model\footnote{Since GPT-OSS-20B itself performs better than Qwen3-30B without extraction, we consider it the stronger of the two smaller models, consistent with the observed trends, despite having fewer parameters.} leads to substantial improvements. For example, when the reasoning model is GPT-OSS-20B, pairing it with the stronger Qwen3-480B extractor yields the best performance, reaching 32.16\% / 29.25\%, compared to only 25.26\% / 20.18\% without extraction. Similarly, when the reasoning model is Qwen3-30B, the 480B extractor provides a gain of nearly 6 absolute points on both metrics.  
Interestingly, the effect is asymmetric: a weak reasoner benefits strongly from a powerful extractor, but a strong reasoner gains little from a weak extractor. For instance, GPT-OSS-20B with a Qwen3-30B extractor (25.71\% / 20.83\%) is close to its no-extraction baseline, suggesting that noisy structural cues are largely ignored or overridden by the reasoning model’s own analysis. In contrast, Qwen3-30B paired with the 480B extractor enjoys a substantial jump, confirming that accurate structural sketches act as an external scaffold that smaller models can leverage.  

\subsubsection{Efficiency Analysis}

We further analyze the computational efficiency of our framework in terms of runtime, and the number of agent calls. 
Across spreadsheets, our method requires 97.49 seconds on average for end-to-end extraction, with 19.54 agent calls per spreadsheet. 
The maximum observed memory usage is 21 GB.
These results indicate that the computational cost of our framework is mainly reflected in iterative agent interactions rather than memory consumption. 
This suggests that the proposed extraction-verification pipeline is efficient enough for real-world spreadsheet understanding scenarios, while remaining independent of the specific query content.

\section{Conclusion} \label{sec:con}

In this paper, we presented \name, a two-stage multi-agent framework for spreadsheet understanding. By incrementally exploring localized regions and constructing structural sketches, \name preserves layout semantics while overcoming context-length limitations. With the integration of code execution, image conversion, and \LaTeX{} conversion, the framework effectively captures numerical, visual, and structural cues.
Experiments on SpreadsheetBench show that \name achieves significant gains over prior methods, and ablation studies verify the necessity of verification, structured sketching, and multi-format tools. Future work includes exploring adaptive strategies for tool invocation and region selection in real-world applications.

\section*{Limitations}

Although our framework demonstrates promising results in spreadsheet understanding, several limitations remain to be addressed in future work.
First, our approach inherently depends on the capabilities of existing foundation models, such as GLM-4.5V, which simultaneously support multimodal inputs (e.g., images) and function calls. These models provide the technical backbone that enables our multi-agent design. However, such strong and versatile models are still relatively scarce in the open-source community, which limits reproducibility and the potential for wide adoption. Exploring how to adapt the framework to more accessible or specialized open-source models would be an important step toward broader applicability.
Second, this paper does not investigate model distillation, namely transferring the reasoning and conversational traces of large models into smaller ones. A key obstacle is the lack of large-scale, realistic spreadsheet datasets that can support effective distillation and evaluation. Without such resources, it remains challenging to systematically assess whether smaller models can inherit the reasoning capabilities of larger ones. We leave this as an important direction for future work.

\section*{Ethics Statement}

The models utilized in this paper, Qwen3\footnote{\url{https://github.com/QwenLM/Qwen3}}
, GPT-OSS\footnote{\url{https://github.com/openai/gpt-oss}}
, and GLM-4.5V\footnote{\url{https://github.com/zai-org/GLM-V}}
, are open-sourced and licensed strictly for academic research purposes. Their use in this work adheres to the terms specified by the respective licenses, and no commercial deployment or redistribution of these models has been performed. Similarly, the datasets employed in our study, Spreadsheet Bench\footnote{\url{https://github.com/RUCKBReasoning/SpreadsheetBench}}
 and RealHiT Bench\footnote{\url{https://github.com/cspzyy/RealHiTBench}}
, are also made available exclusively for research purposes, and all experiments were conducted in accordance with these intended uses.

\section*{Acknowledgment}

This project is funded in part by Shenzhen Loop Area Institute, by the Centre for Perceptual and Interactive Intelligence (CPII) Ltd under the Innovation and Technology Commission (ITC)’s InnoHK,  in part by NSFC-RGC Project N\_CUHK498/24, and in part by Guangdong Basic and Applied Basic Research Foundation (No. 2023B1515130008, XW).

\bibliography{reference}

\clearpage\appendix\section*{Appendix}

\section{Additional Experiments} \label{sec:app_exp}

\subsection{RealHiT Bench} \label{sec:realhit}

\paragraph{Baseline.}
In the RealHiT Bench experiments, we establish two primary categories of baselines. The first category consists of models that directly ingest the \LaTeX{} tables as plain text inputs. Since the tables in this benchmark are relatively small, such models can process the entire structure in a single pass without exceeding context limits. However, this approach fails to capture the strengths of step-wise exploration and overlooks many structural and stylistic cues that become crucial in more complex, real-world spreadsheets. The second category of baselines involves models that operate over file inputs and attempt step-by-step reasoning, simulating an incremental reading process. To strengthen the comparison, we also include TreeThinker as a representative method for structured chain-of-thought reasoning. 

\paragraph{LaTeX input.}  
As shown in Table~\ref{tab:realhit_latex}, when RealHiT tables are directly provided in \LaTeX{} format, both TreeThinker and \name offer noticeable improvements on smaller models, while the benefits diminish for stronger backbones. For example, on Llama3.1-8B, \name increases Fact Checking EM from 33.20 to 34.51 and Numerical Reasoning EM from 10.38 to 15.69, outperforming TreeThinker in all three sub-tasks. A similar trend is observed on Qwen2.5-7B, where \name achieves 51.27 EM in Fact Checking and 32.30 EM in Numerical Reasoning, surpassing both the plain re-evaluated baseline and TreeThinker. However, for larger models such as Llama3.3-70B and Qwen2.5-72B, the improvements become marginal. For instance, \name only yields a +0.65 EM gain on Numerical Reasoning with Qwen2.5-72B, while showing slight regressions on Structure Comprehending. These results suggest that when the backbone itself is sufficiently strong, direct \LaTeX{} input already enables effective reasoning, leaving limited headroom for additional exploration mechanisms.

\paragraph{File input.}  
In contrast, as shown in Table~\ref{tab:realhit_python}, when feeding models with spreadsheet files and enabling incremental exploration, both TreeThinker and \name bring consistent gains across all backbones, and the improvements from \name are substantially larger. For example, on GPT-OSS-20B, \name raises Fact Checking EM by +5.92 and Numerical Reasoning EM by +9.99 compared to the plain baseline, outperforming TreeThinker by wide margins. On Qwen3-30B, \name delivers even more pronounced benefits, improving Fact Checking from 37.63 (TreeThinker) to 46.18 EM and Numerical Reasoning from 31.26 to 35.80 EM. Importantly, these advantages also extend to very large models: with Qwen3-Coder-480B, \name reaches 69.10 EM / 77.18 F1 in Fact Checking and 76.26 EM / 82.71 F1 in Structure Comprehending, while TreeThinker suffers severe performance drops. These findings confirm that step-by-step file exploration with multi-format inputs is crucial, as it provides systematic gains across scales and tasks, demonstrating that \name can robustly handle complex spreadsheet reasoning beyond the limits of text-only representations.

\subsection{Case Study}

\begin{figure*}[t]
    \centering
    \includegraphics[width=\textwidth]{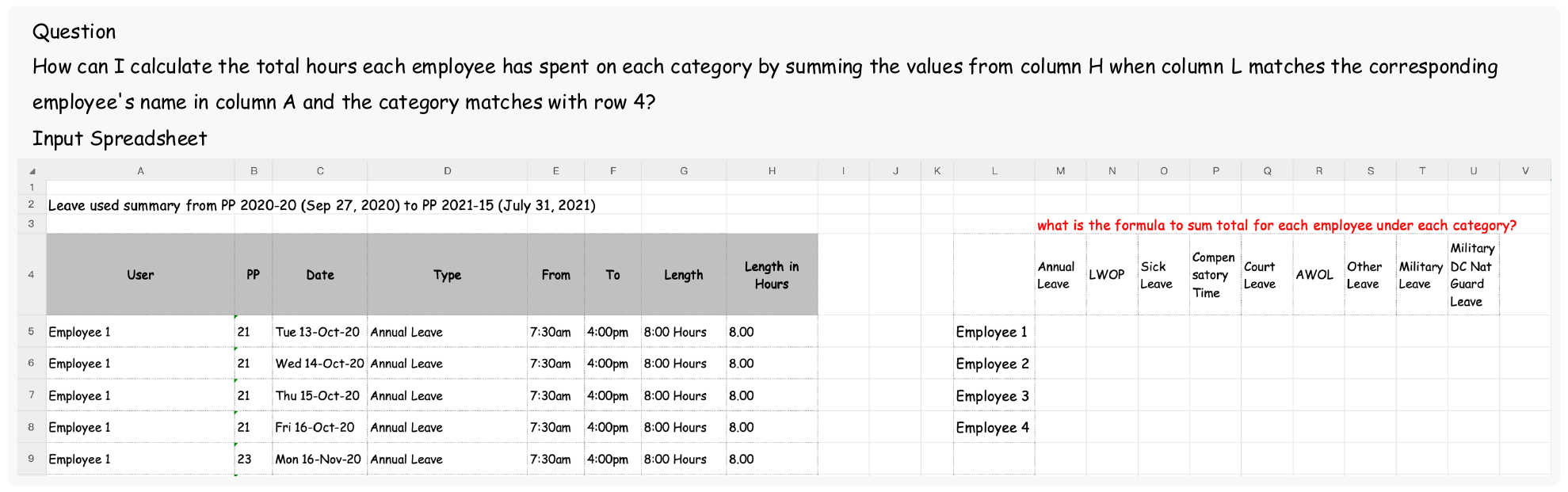}
    \caption{Case 1: Partial view of the leave entries spreadsheet.}
    \label{fig:case_1}
\end{figure*}

\begin{figure*}[t]
    \centering
    \includegraphics[width=0.7\textwidth]{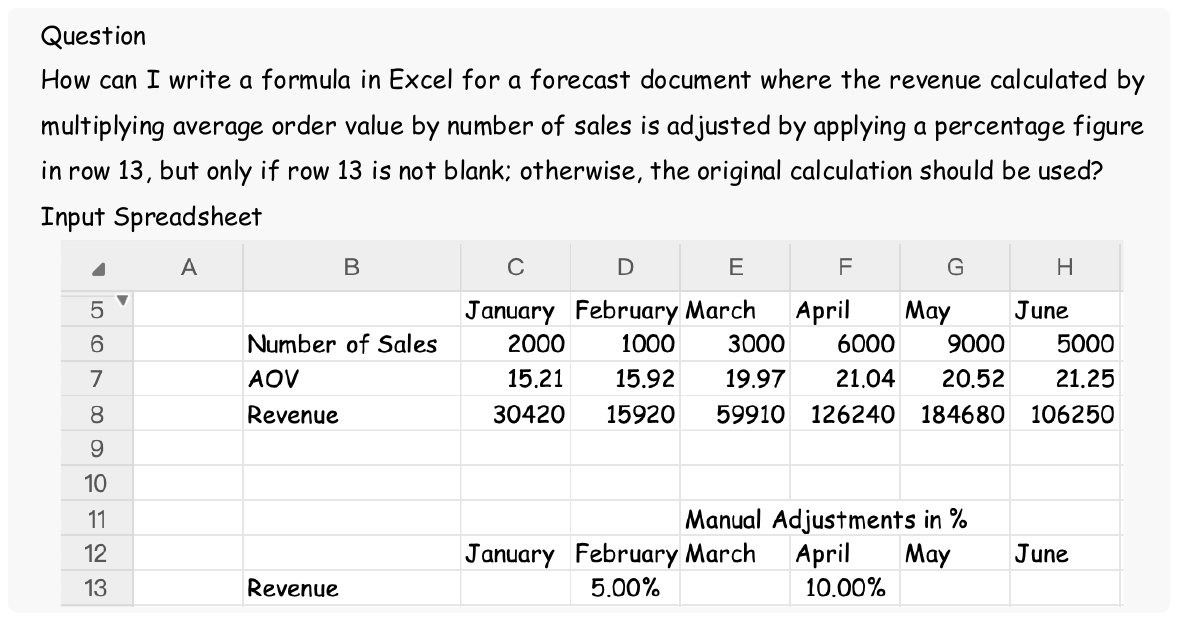}
    \caption{Case 2: Partial view of the sales data spreadsheet.}
    \label{fig:case_2}
\end{figure*}


Here, we present a representative case to illustrate the effectiveness of the proposed method, shown in Figure~\ref{fig:case_1}. The generated codes from both the baseline method and our proposed method are presented in Table~\ref{tab:case_1}.
In this case, the baseline method failed because it mistakenly applied the employee filter to column $L$ of the detail rows, rather than the correct column $A$ (\texttt{User}). Since column $L$ in the detail area is mostly empty, almost all rows were ignored, and the totals degenerated into trivial values close to zero. A snippet of the problematic code is shown below:
\begin{lstlisting}
# BUG: employee filter is applied to col L (index 11)
df.iloc[i, 11] == employee

# Correct condition should be:
df.iloc[i, 0] == employee
\end{lstlisting}
This error highlights a fundamental limitation of text-only approaches: without explicit structural awareness, the model cannot reliably distinguish between header rows, detail areas, and summary regions. As a result, it may select the wrong reference column or misalign the header–data correspondence.

By contrast, with structure extraction, our method explicitly recognizes that the spreadsheet contains two semantically distinct regions:
\begin{lstlisting}
### Main Table: Leave Entries
```yaml
...

### Designated Summary Area
```yaml
...
\end{lstlisting}
The exploration agent identifies the main data table (leave entries) as the core source of employee-level records, while also detecting the designated summary area as a separate, higher-level aggregation. This structural sketch prevents the misinterpretation of empty or auxiliary columns as valid keys. Moreover, because our framework incorporates layout cues and region-based parsing, the model correctly aligns headers with the corresponding data fields. It then applies the filter condition to column $A$ (\texttt{User}), thereby including all valid rows and yielding accurate totals.

Another case is shown in Figure~\ref{fig:case_2}. The generated codes from both the baseline method and our proposed method are presented in Table~\ref{tab:case_2}.
In this case, the baseline method failed because it mistakenly wrote the computed answer into line 9, which is outside the actual table boundary. A snippet of the problematic code is shown below:
\begin{lstlisting}
df.iloc [8, col] = adjusted_revenue
\end{lstlisting}
This error occurred because the baseline approach lacked an explicit understanding of the spreadsheet’s structural boundaries.

By contrast, our method leverages structure extraction to explicitly recognize the table boundaries. In this example, the exploration agent identified that the core data table is constrained to the range B5:H8 (noting that spreadsheet notation starts at 1, whereas arrays are 0-indexed). This structured annotation is recorded as:
\begin{lstlisting}
  node_3:
    start_index: "B8"
    end_index: "B8"
    value: "Revenue"
    children: []
\end{lstlisting}
With this structured context, the model becomes explicitly aware of the target location. It understands that the adjusted revenue should be placed in row 8, and any attempt to place it outside this range would lead to an inconsistency with the spreadsheet layout. As a result, the model ensures that the answer is correctly filled into the designated cell in row 8, avoiding any misplacement.

\begin{table*}[t]
  \centering
  \caption{Generated codes for Case 1 (leave entries).}
\begin{lstlisting}
# Code Generated by Qwen3-Coder-480B
import pandas as pd
from openpyxl import load_workbook

df = pd.read_excel('/mnt/data/input/1_51859_input.xlsx', header=None)

employees = ['Employee 1', 'Employee 2', 'Employee 3', 'Employee 4']

categories = []
for i in range(12, min(21, len(df.columns))):
    if pd.notna(df.iloc[3, i]):
        categories.append(df.iloc[3, i])
results = {}

for employee in employees:
    results[employee] = {}
    for category in categories:
        total_hours = 0
        for i in range(4, len(df)):
            if (pd.notna(df.iloc[i, 11]) and 
                df.iloc[i, 11] == employee and 
                pd.notna(df.iloc[i, 3]) and 
                df.iloc[i, 3] == category and
                pd.notna(df.iloc[i, 7])):
                total_hours += df.iloc[i, 7]
        results[employee][category] = total_hours

wb = load_workbook('/mnt/data/input/1_51859_input.xlsx')
ws = wb.active

for i, employee in enumerate(employees):
    row_index = 5 + i
    for j, category in enumerate(categories):
        col_index = 13 + j
        ws.cell(row=row_index, column=col_index, value=results[employee][category])

wb.save('/mnt/data/output/1_51859_output.xlsx')
\end{lstlisting}
\rule{\textwidth}{0.5pt}
\begin{lstlisting}
# Code Generated by Qwen3-Coder-480B w/ StructureAgent
import pandas as pd

file_path = "/mnt/data/input/1_51859_input.xlsx"
df = pd.read_excel(file_path, sheet_name="Leave used by employee", header=None)

data_rows = df.iloc[4:194, [0, 3, 7]]
data_rows.columns = ['User', 'Type', 'Hours']

data_rows['Hours'] = pd.to_numeric(data_rows['Hours'], errors='coerce').fillna(0)

expected_employees = df.iloc[4:8, 11].dropna().tolist()
expected_leave_types = df.iloc[3, 12:21].tolist() 

summary_table = pd.pivot_table(data_rows, values='Hours', index='User', columns='Type', aggfunc='sum', fill_value=0)

summary_table = summary_table.reindex(index=expected_employees, columns=expected_leave_types, fill_value=0)

for i, employee in enumerate(expected_employees):
    for j, leave_type in enumerate(expected_leave_types):
        row_idx = 4 + i
        col_idx = 12 + j
        df.iloc[row_idx, col_idx] = summary_table.loc[employee, leave_type]

output_path = "/mnt/data/output/1_51859_output.xlsx"
df.to_excel(output_path, sheet_name="Leave used by employee", header=False)
\end{lstlisting}
\label{tab:case_1}
\end{table*}

\begin{table*}[t]
  \centering
  \caption{Generated codes for Case 2 (sales data).}
\begin{lstlisting}
# Code Generated by Qwen3-Coder-480B
import pandas as pd

df = pd.read_excel('/mnt/data/input/1_52888_input.xlsx', header=None)

for col in range(2, 8):
    num_sales = df.iloc[5, col]
    aov = df.iloc[6, col]
    adjustment = df.iloc[12, col]
    
    base_revenue = num_sales * aov
    
    if pd.notna(adjustment):
        adjusted_revenue = base_revenue * (1 + adjustment)
    else:
        adjusted_revenue = base_revenue
        
    df.iloc[8, col] = adjusted_revenue

df.to_excel('/mnt/data/output/1_52888_output.xlsx', index=False, header=False)
\end{lstlisting}
\rule{\textwidth}{0.5pt}
\begin{lstlisting}
# Code Generated by Qwen3-Coder-480B w/ StructureAgent
import pandas as pd

df = pd.read_excel('/mnt/data/input/1_52888_input.xlsx', header=None)

number_of_sales = df.iloc[5, 2:8].values
aov = df.iloc[6, 2:8].values

base_revenue = number_of_sales * aov

adjustments = df.iloc[12, 2:8].values

adjusted_revenue = []
for i in range(len(base_revenue)):
    if pd.notna(adjustments[i]) and adjustments[i] != 0:
        adjusted_value = base_revenue[i] * (1 + adjustments[i])
        adjusted_revenue.append(adjusted_value)
    else:
        adjusted_revenue.append(base_revenue[i])

df.iloc[7, 2:8] = adjusted_revenue

df.to_excel('/mnt/data/output/1_52888_output.xlsx', sheet_name='Sheet1', index=False, header=False)
\end{lstlisting}
\label{tab:case_2}
\end{table*}
\begin{table*}[t]
\centering
\caption{Performance comparison across different models on three sub-tasks of RealHiT Bench: Fact Checking, Numerical Reasoning, and Structure Comprehending. Results marked with $^*$ indicate re-evaluations under a slightly modified prompt that encourages models to perform explicit reasoning before producing final answers, as opposed to directly outputting results.}
\setlength{\tabcolsep}{5pt}
\begin{tabular}{l|cc|cc|cc}
\toprule
\multirow{2.5}{*}{Model}  & \multicolumn{2}{c|}{Fact Checking} & \multicolumn{2}{c|}{Numerical Reasoning} & \multicolumn{2}{c}{Structure Comprehending} \\ \cmidrule(lr){2-7}
 & EM & F1 & EM & F1 & EM & F1 \\
\midrule
GPT4o & 60.31 & 68.97 & 38.65 & 50.12 & 63.04 & 71.14 \\
Gemini1.5-pro & 59.08 & 66.14 & 35.54 & 43.74 & 63.64 & 69.71 \\
DeepSeek-R1 & 70.91 & 79.45 & 70.31 & 72.54 & 82.71 & 84.62 \\
\midrule
Llama3.1-8B-Instruct & 30.32 & 44.93 & 14.53 & 27.21 & 35.90 & 50.80 \\
Llama3.1-8B-Instruct$^*$ & 33.20 & 44.15 & 10.38 & 22.58 & 25.51 & 39.51 \\
\quad w/ TreeThinker & 32.37 & 44.93 & 13.36 & 24.76 & 21.72 & 38.78 \\
\quad w/ \name & \textbf{34.51} & \textbf{47.68} & \textbf{15.69} & \textbf{28.36} & \textbf{25.25} & \textbf{40.33} \\
\midrule
Qwen2.5-7B-Instruct & 18.65 & 38.39 & 5.32 & 19.75 & 23.48 & 44.81 \\
Qwen2.5-7B-Instruct$^*$ & 48.56 & 54.70 & 29.83 & 38.29 & 45.96 & 55.45 \\
\quad w/ TreeThinker & 51.19 & 57.45 & 31.13 & 39.14 & 44.95 & 53.63 \\
\quad w/ \name & \textbf{51.27} & \textbf{57.56} & \textbf{32.30} & \textbf{40.44} & \textbf{48.74} & \textbf{56.61} \\
\midrule
Llama3.3-70B-Instruct & 53.08 & 64.53 & 36.58 & 48.99 & 55.81 & 68.93 \\
Llama3.3-70B-Instruct$^*$ & 67.13 & \textbf{73.14} & 54.47 & 61.60 & 73.99 & 78.61 \\
\quad w/ TreeThinker & \textbf{67.30} & 73.06 & 57.98 & \textbf{64.92} & \textbf{74.24} & \textbf{79.49} \\
\quad w/ \name & 65.74 & 71.94 & \textbf{58.37} & 64.44 & \textbf{74.24} & 79.36 \\
\midrule
Qwen2.5-72B-Instruct & 51.93 & 62.15 & 26.98 & 39.23 & 54.55 & 68.34 \\
Qwen2.5-72B-Instruct$^*$ & 65.74 & 72.93 & 52.40 & 61.33 & \textbf{73.48} & \textbf{80.54} \\
\quad w/ TreeThinker & \textbf{66.80} & \textbf{74.03} & 50.97 & 60.33 & 72.73 & 79.45 \\
\quad w/ \name & 66.64 & 73.44 & \textbf{53.05} & \textbf{62.31} & 70.71 & 78.42 \\
\bottomrule
\end{tabular}
\label{tab:realhit_latex}
\end{table*}

\begin{table*}[t]
\centering
\caption{Performance comparison across different models with Python tool on three sub-tasks of RealHiT Bench: Fact Checking, Numerical Reasoning, and Structure Comprehending.}
\setlength{\tabcolsep}{5pt}
\begin{tabular}{l|cc|cc|cc}
\toprule
\multirow{2.5}{*}{Model}  & \multicolumn{2}{c|}{Fact Checking} & \multicolumn{2}{c|}{Numerical Reasoning} & \multicolumn{2}{c}{Structure Comprehending} \\ \cmidrule(lr){2-7}
 & EM & F1 & EM & F1 & EM & F1 \\
\midrule
GPT-OSS-20B & 42.81 & 48.54 & 34.63 & 39.03 & 33.84 & 40.46 \\
\quad w/ TreeThinker & 45.11 & 50.12 & 39.82 & 44.42 & 33.59 & 39.53 \\
\quad w/ \name & \textbf{48.73} & \textbf{53.64} & \textbf{44.62} & \textbf{48.92} & \textbf{39.65} & \textbf{44.74} \\
\midrule
GPT-OSS-120B & 58.18 & 62.75 & 55.38 & 59.16 & 54.04 & 57.19 \\
\quad w/ TreeThinker & 56.53 & 61.14 & 54.60 & 59.06 & 53.03 & 57.98 \\
\quad w/ \name & \textbf{64.09} & \textbf{69.22} & \textbf{58.24} & \textbf{62.58} & \textbf{62.37} & \textbf{66.01} \\
\midrule
Qwen3-30B & 36.24 & 42.31 & 28.66 & 35.76 & 37.88 & 44.26 \\
\quad w/ TreeThinker & 37.63 & 45.03 & 31.26 & 39.77 & 34.09 & 41.65 \\
\quad w/ \name & \textbf{46.18} & \textbf{52.44} & \textbf{35.80} & \textbf{44.36} & \textbf{40.15} & \textbf{47.28} \\
\midrule
Qwen3-235B & 56.94 & 62.60 & 42.54 & 48.94 & 61.87 & 67.85 \\
\quad w/ TreeThinker & 57.27 & 62.54 & 44.10 & 50.06 & 59.09 & 63.28 \\
\quad w/ \name & \textbf{62.70} & \textbf{68.55} & \textbf{47.47} & \textbf{54.25} & \textbf{63.89} & \textbf{69.43} \\
\midrule
Qwen3-Coder-480B & 66.47 & 74.46 & 54.60 & 64.97 & 72.98 & 80.01 \\
\quad w/ TreeThinker & 57.85 & 65.87 & 47.21 & 58.05 & 58.33 & 65.77 \\
\quad w/ \name & \textbf{69.10} & \textbf{77.18} & \textbf{55.25} & \textbf{65.74} & \textbf{76.26} & \textbf{82.71} \\
\bottomrule
\end{tabular}
\label{tab:realhit_python}
\end{table*}

\clearpage

\begin{tcolorbox}[
    breakable,
    float*=h,
    width=\textwidth, 
    colback=background, 
    colframe=title, 
    fonttitle=\bfseries, 
    title=Prompt for Structure Extraction
]
\text{}You are a spreadsheet expert who can analyze and manipulate spreadsheets using Python. Your task is to detect tables in a spreadsheet and generate hierarchical structures for their top-row and left-column headers.   \\ 
\text{~}\\ 
\text{}(1) Detect All Tables \\ 
\text{~~}Identify all rectangular areas in the spreadsheet that can be considered tables. A table is defined as a block with identifiable headers and corresponding data. Clarify how to handle: \\ 
\text{~~}- Empty rows/columns (skip or stop). \\ 
\text{~~}- Multiple adjacent tables (detect separately). \\ 
\text{~~}- Merged cells (treat as higher-level headers). \\ 
\text{~}\\ 
\text{}(2) Analyze and Output Table Structure \\ 
\text{~}\\ 
\text{}\#\#\#\# a. Metadata \\ 
\text{}- Sheet Name: The sheet containing the table.   \\ 
\text{}- Table Name: If defined (e.g., Table1 in Excel).   \\ 
\text{}- Table Range: The full rectangular range of the table (e.g., A1:D20).   \\ 
\text{}- Data Range: The rectangular range containing only data (excluding headers).   \\ 
\text{}- Notes / Footnotes: Optional annotations or data source lines below the table.  \\ 
\text{~}\\ 
\text{}\#\#\#\# b. Header Structure \\ 
\text{}- Determine Header Format \\ 
\text{~~}First, classify the table structure into one of the following:   \\ 
\text{~~}- Column-only header (most common: header row(s) at the top)   \\ 
\text{~~}- Row-only header (rare: header column(s) on the left)   \\ 
\text{~~}- Both row \& column headers (matrix-style tables)   \\ 
\text{~}\\ 
\text{}- Detect all header cell:   \\ 
\text{~~}Each header cell have three attributes: start position, end position, and value.   \\ 
\text{~~}- Start Position: The coordinate of the top-left cell of the header (e.g., A1).   \\ 
\text{~~}- End Position: The coordinate of the bottom-right cell of the header (e.g., B2).   \\ 
\text{~~}- Value: The text content of the header cell. \\ 
\text{~}\\ 
\text{}- Group header cells with the same level: \\ 
\text{~~}Headers can be nested, indicating hierarchical relationships. Group headers based on their indentation levels or merged cell spans. For example, if "Country" spans two columns and "State" and "City" are under it, they should be grouped accordingly. \\ 
\text{~}\\ 
\text{}- Construct hierarchical structures: \\ 
\text{~~}Build the hierarchical structure for both row and column headers. Each node in the hierarchy should include: \\ 
\text{~~}- Start Index: The starting cell coordinate of the header. \\ 
\text{~~}- End Index: The ending cell coordinate of the header. \\ 
\text{~~}- Value: The text content of the header. \\ 
\text{~~}- Children: A list of child nodes representing sub-headers. \\ 
\text{~}\\ 
\text{}\#\#\#\# c. Data Properties \\ 
\text{}For each column (or row header if applicable), detect:   \\ 
\text{}- Data Type (string, number, date, boolean, mixed)   \\ 
\text{}- Unit (if explicit, e.g., km, dollar) \\ 
\text{}- Format (e.g., bold, italic, background color, currency)   \\ 
\text{~}\\ 
\text{}(3) Output Format \\ 
\text{}For each table, please generate a hierarchical structure in the following YAML format (with your reasoning content and other summary infomation). \\ 
\text{}\textasciigrave\textasciigrave\textasciigrave yaml \\ 
\text{}sheet\_name: sheet\_name \\ 
\text{}table\_name: table\_name \\ 
\text{}table\_range: table\_range \\ 
\text{}data\_range: data\_range \\ 
\text{}notes: \\ 
\text{~~}- ... \\ 
\text{~~}- ... \\ 
\text{}row\_header: \\ 
\text{~~~~}node\_1: \\ 
\text{~~~~~~~~}start\_index: A3 \\ 
\text{~~~~~~~~}end\_index: A4 \\ 
\text{~~~~~~~~}value: City \\ 
\text{~~~~~~~~}children: \\ 
\text{~~~~~~~~~~~~}- node\_1.1: \\ 
\text{~~~~~~~~~~~~~~~~}start\_index: B3 \\ 
\text{~~~~~~~~~~~~~~~~}end\_index: B3 \\ 
\text{~~~~~~~~~~~~~~~~}value: Population \\ 
\text{~~~~~~~~~~~~~~~~}children: ... \\ 
\text{~~~~~~~~~~~~}- node\_1.2: \\ 
\text{~~~~~~~~~~~~~~~~}start\_index: B4 \\ 
\text{~~~~~~~~~~~~~~~~}end\_index: B4 \\ 
\text{~~~~~~~~~~~~~~~~}value: Area \\ 
\text{~~~~~~~~~~~~~~~~}children: ... \\ 
\text{}column\_header: \\ 
\text{~~~~}node\_1: \\ 
\text{~~~~~~~~}start\_index: C1 \\ 
\text{~~~~~~~~}end\_index: D1 \\ 
\text{~~~~~~~~}value: Country \\ 
\text{~~~~~~~~}children: \\ 
\text{~~~~~~~~~~~~}- node\_1.1: \\ 
\text{~~~~~~~~~~~~~~~~}start\_index: C2 \\ 
\text{~~~~~~~~~~~~~~~~}end\_index: C2 \\ 
\text{~~~~~~~~~~~~~~~~}value: State \\ 
\text{~~~~~~~~~~~~~~~~}children: ... \\ 
\text{~~~~~~~~~~~~}- node\_1.2: \\ 
\text{~~~~~~~~~~~~~~~~}start\_index: D2 \\ 
\text{~~~~~~~~~~~~~~~~}end\_index: D2 \\ 
\text{~~~~~~~~~~~~~~~~}value: City \\ 
\text{~~~~~~~~~~~~~~~~}children: ... \\ 
\text{}data\_properties: \\ 
\text{~~}Population: \\ 
\text{~~~~}type: number \\ 
\text{~~~~}unit: people \\ 
\text{~~~~}format: plain \\ 
\text{~~}Area: \\ 
\text{~~~~}type: number \\ 
\text{~~~~}unit: km \\ 
\text{~~~~}format: comma-separated \\ 
\text{}\textasciigrave\textasciigrave\textasciigrave \\ 
\text{~}\\ 
\text{}\#\#\# spreadsheet\_path \\ 
\text{}\textcolor{blue}{\{spreadsheet\_path\}} \\ 
\text{~}\\ 
\text{}\#\#\# sheet\_name \\ 
\text{}\textcolor{blue}{\{sheet\_name\}} \\ 
\text{~}\\ 
\text{}\#\#\# used\_range \\ 
\text{}\textcolor{blue}{\{used\_range\}}
\end{tcolorbox}

\begin{tcolorbox}[
    breakable,
    float*=h,
    width=\textwidth, 
    colback=background, 
    colframe=title, 
    fonttitle=\bfseries, 
    title=Prompt for Verification
]
\text{}You are a Spreadsheet Structure Verification Assistant. Your task is to verify whether a spreadsheet's structure matches the expected description provided.  \\ 
\text{~}\\ 
\text{}\#\#\# Instructions: \\ 
\text{}1. Utilize the tool to get the information and determine if the structure is **correct and consistent**.   \\ 
\text{}2. If there are discrepancies (e.g., mismatched ranges, missing/extra rows or columns, incorrect headers, or formatting inconsistencies), describe them clearly and precisely.   \\ 
\text{}3. If the structure is valid, confirm that no issues were found.   \\ 
\text{~}\\ 
\text{}\#\#\# Output Format: \\ 
\text{}\textasciigrave\textasciigrave\textasciigrave yaml \\ 
\text{}verification: true/false, \\ 
\text{}issues: \\ 
\text{~~~~}- "Description of issue 1", \\ 
\text{~~~~}- "Description of issue 2" \\ 
\text{}\textasciigrave\textasciigrave\textasciigrave \\ 
\text{}- verification is true: Structure is correct, "issues" should be an empty list. \\ 
\text{}- verification is false: One or more issues found, list them in "issues". \\ 
\text{~}\\ 
\text{}\#\# Information to Verify: \\ 
\text{~}\\ 
\text{}- Spreadsheet Path: \textcolor{blue}{\{spreadsheet\_path\}} \\ 
\text{}- Sheet Name: \textcolor{blue}{\{sheet\_name\}} \\ 
\text{}- Used Range: \textcolor{blue}{\{used\_range\}} \\ 
\text{}- Spreadsheet Structure: \\ 
\text{}\textcolor{blue}{\{spreadsheet\_info\}}

\end{tcolorbox}

\begin{tcolorbox}[
    breakable,
    float*=t,
    width=\textwidth, 
    colback=background, 
    colframe=title, 
    fonttitle=\bfseries, 
    title=Definition of Code Execution Tool
]
\text{}\{ \\
\text{~~~~}"type": "function", \\
\text{~~~~}"function": \{ \\
\text{~~~~~~~~}"name": "execute\_python", \\
\text{~~~~~~~~}"description": "When you send a message containing Python code to python, it will be executed in a stateful Jupyter notebook environment. python will respond with the output of the execution or time out after 60.0 seconds. The drive at '/mnt/data' can be used to save and persist user files. Internet access for this session is disabled.", \\
\text{~~~~~~~~}"parameters": \{ \\
\text{~~~~~~~~~~~~}"type": "object", \\
\text{~~~~~~~~~~~~}"properties": \{ \\
\text{~~~~~~~~~~~~~~~~}"code": \{ \\
\text{~~~~~~~~~~~~~~~~~~~~}"type": "string", \\
\text{~~~~~~~~~~~~~~~~~~~~}"description": "The Python code to execute", \\
\text{~~~~~~~~~~~~~~~~}\} \\
\text{~~~~~~~~~~~~}\}, \\
\text{~~~~~~~~~~~~}"required": ["code"] \\
\text{~~~~~~~~}\}, \\
\text{~~~~}\} \\
\text{}\} \\
\end{tcolorbox}

\begin{tcolorbox}[
    breakable,
    float*=t,
    width=\textwidth, 
    colback=background, 
    colframe=title, 
    fonttitle=\bfseries, 
    title=Definition of Vision Range Description Tool
]
\text{}\{ \\
\text{~~~~}"type": "function", \\
\text{~~~~}"function": \{ \\
\text{~~~~~~~~}"name": "vision\_question\_answer", \\
\text{~~~~~~~~}"description": "You can analyze a selected range of cells in a spreadsheet using vision capabilities. When asking, focus on small, well-defined questions and describe all relevant details. Don't just hand over your entire task to the tool.", \\
\text{~~~~~~~~}"parameters": \{ \\
\text{~~~~~~~~~~~~}"type": "object", \\
\text{~~~~~~~~~~~~}"properties": \{ \\
\text{~~~~~~~~~~~~~~~~}"path": \{ \\
\text{~~~~~~~~~~~~~~~~~~~~}"type": "string", \\
\text{~~~~~~~~~~~~~~~~~~~~}"description": "The file path to the spreadsheet containing the image." \\
\text{~~~~~~~~~~~~~~~~}\}, \\
\text{~~~~~~~~~~~~~~~~}"sheet\_name": \{ \\
\text{~~~~~~~~~~~~~~~~~~~~}"type": "string", \\
\text{~~~~~~~~~~~~~~~~~~~~}"description": "The name of the sheet from which to extract the image." \\
\text{~~~~~~~~~~~~~~~~}\}, \\
\text{~~~~~~~~~~~~~~~~}"range": \{ \\
\text{~~~~~~~~~~~~~~~~~~~~}"type": "string", \\
\text{~~~~~~~~~~~~~~~~~~~~}"description": "Optional. The cell range to search for the image (e.g., 'A1:D20'). Use to limit extraction to a specific area. Images smaller than 8192×16384 pixels are supported, so keep the range within reasonable bounds." \\
\text{~~~~~~~~~~~~~~~~}\}, \\
\text{~~~~~~~~~~~~~~~~}"question": \{ \\
\text{~~~~~~~~~~~~~~~~~~~~}"type": "string", \\
\text{~~~~~~~~~~~~~~~~~~~~}"description": "The question to answer about the specified range of cells." \\
\text{~~~~~~~~~~~~~~~~}\} \\
\text{~~~~~~~~~~~~}\}, \\
\text{~~~~~~~~~~~~}"required": ["path", "sheet\_name", "range", "question"] \\
\text{~~~~~~~~}\} \\
\text{~~~~}\} \\
\text{}\} \\
\end{tcolorbox}

\begin{tcolorbox}[
    breakable,
    float*=h,
    width=\textwidth, 
    colback=background, 
    colframe=title, 
    fonttitle=\bfseries, 
    title=Definition of Latex Range Description Tool
]
\text{}\{ \\
\text{~~~~}"type": "function", \\
\text{~~~~}"function": \{ \\
\text{~~~~~~~~}"name": "latex\_question\_answer", \\
\text{~~~~~~~~}"description": "You can analyze a selected range of cells in a spreadsheet using latex capabilities. When asking, focus on small, well-defined questions and describe all relevant details — don't just hand over your entire task to the tool.", \\
\text{~~~~~~~~}"parameters": \{ \\
\text{~~~~~~~~~~~~}"type": "object", \\
\text{~~~~~~~~~~~~}"properties": \{ \\
\text{~~~~~~~~~~~~~~~~}"path": \{ \\
\text{~~~~~~~~~~~~~~~~~~~~}"type": "string", \\
\text{~~~~~~~~~~~~~~~~~~~~}"description": "The file path to the Excel file." \\
\text{~~~~~~~~~~~~~~~~}\}, \\
\text{~~~~~~~~~~~~~~~~}"sheet\_name": \{ \\
\text{~~~~~~~~~~~~~~~~~~~~}"type": "string", \\
\text{~~~~~~~~~~~~~~~~~~~~}"description": "The name of the sheet from which to extract the latex." \\
\text{~~~~~~~~~~~~~~~~}\}, \\
\text{~~~~~~~~~~~~~~~~}"range": \{ \\
\text{~~~~~~~~~~~~~~~~~~~~}"type": "string", \\
\text{~~~~~~~~~~~~~~~~~~~~}"description": "Optional. The cell range to search for the latex (e.g., 'A1:D20'). Use to limit extraction to a specific area." \\
\text{~~~~~~~~~~~~~~~~}\}, \\
\text{~~~~~~~~~~~~~~~~}"question": \{ \\
\text{~~~~~~~~~~~~~~~~~~~~}"type": "string", \\
\text{~~~~~~~~~~~~~~~~~~~~}"description": "The question to answer about the specified range of cells." \\
\text{~~~~~~~~~~~~~~~~}\} \\
\text{~~~~~~~~~~~~}\}, \\
\text{~~~~~~~~~~~~}"required": ["path", "sheet\_name", "range", "question"] \\
\text{~~~~~~~~}\} \\
\text{~~~~}\} \\
\text{}\} \\
\end{tcolorbox}

\begin{tcolorbox}[
    breakable,
    float*=h,
    width=\textwidth, 
    colback=background, 
    colframe=title, 
    fonttitle=\bfseries, 
    title=Definition of Excel to Vision Tool
]
\text{}\{ \\
\text{~~~~}"type": "function", \\
\text{~~~~}"function": \{ \\
\text{~~~~~~~~}"name": "convert\_excel\_to\_image", \\
\text{~~~~~~~~}"description": "Extracts an image from a specified spreadsheet file.", \\
\text{~~~~~~~~}"parameters": \{ \\
\text{~~~~~~~~~~~~}"type": "object", \\
\text{~~~~~~~~~~~~}"properties": \{ \\
\text{~~~~~~~~~~~~~~~~}"path": \{ \\
\text{~~~~~~~~~~~~~~~~~~~~}"type": "string", \\
\text{~~~~~~~~~~~~~~~~~~~~}"description": "The file path to the Excel file." \\
\text{~~~~~~~~~~~~~~~~}\}, \\
\text{~~~~~~~~~~~~~~~~}"sheet\_name": \{ \\
\text{~~~~~~~~~~~~~~~~~~~~}"type": "string", \\
\text{~~~~~~~~~~~~~~~~~~~~}"description": "The name of the sheet from which to extract the image." \\
\text{~~~~~~~~~~~~~~~~}\}, \\
\text{~~~~~~~~~~~~~~~~}"range": \{ \\
\text{~~~~~~~~~~~~~~~~~~~~}"type": "string", \\
\text{~~~~~~~~~~~~~~~~~~~~}"description": "The cell range to search for the image (e.g., 'A1:D20'). Use to limit extraction to a specific area. Images smaller than 8192×16384 pixels are supported, so keep the range within reasonable bounds." \\
\text{~~~~~~~~~~~~~~~~}\} \\
\text{~~~~~~~~~~~~}\}, \\
\text{~~~~~~~~~~~~}"required": ["path", "sheet\_name", "range"] \\
\text{~~~~~~~~}\} \\
\text{~~~~}\} \\
\text{}\} \\
\end{tcolorbox}

\begin{tcolorbox}[
    breakable,
    float*=h,
    width=\textwidth, 
    colback=background, 
    colframe=title, 
    fonttitle=\bfseries, 
    title=Definition of Excel to Latex Tool
]
\text{}\{ \\
\text{~~~~}"type": "function", \\
\text{~~~~}"function": \{ \\
\text{~~~~~~~~}"name": "convert\_excel\_to\_latex", \\
\text{~~~~~~~~}"description": "Converts an Excel file to a LaTeX file.", \\
\text{~~~~~~~~}"parameters": \{ \\
\text{~~~~~~~~~~~~}"type": "object", \\
\text{~~~~~~~~~~~~}"properties": \{ \\
\text{~~~~~~~~~~~~~~~~}"path": \{ \\
\text{~~~~~~~~~~~~~~~~~~~~}"type": "string", \\
\text{~~~~~~~~~~~~~~~~~~~~}"description": "The file path to the Excel file." \\
\text{~~~~~~~~~~~~~~~~}\}, \\
\text{~~~~~~~~~~~~~~~~}"sheet\_name": \{ \\
\text{~~~~~~~~~~~~~~~~~~~~}"type": "string", \\
\text{~~~~~~~~~~~~~~~~~~~~}"description": "The name of the sheet from which to extract the image." \\
\text{~~~~~~~~~~~~~~~~}\}, \\
\text{~~~~~~~~~~~~~~~~}"range": \{ \\
\text{~~~~~~~~~~~~~~~~~~~~}"type": "string", \\
\text{~~~~~~~~~~~~~~~~~~~~}"description": "The cell range to convert to LaTeX. Use to limit conversion to a specific area." \\
\text{~~~~~~~~~~~~~~~~}\} \\
\text{~~~~~~~~~~~~}\}, \\
\text{~~~~~~~~~~~~}"required": ["path", "sheet\_name", "range"] \\
\text{~~~~~~~~}\} \\
\text{~~~~}\} \\
\text{}\} \\
\end{tcolorbox}

\begin{tcolorbox}[
    breakable,
    float*=h,
    width=\textwidth, 
    colback=background, 
    colframe=title, 
    fonttitle=\bfseries, 
    title=YAML Structure Example
]
\text{}\#\#\# Main Table: Leave Entries\\
\text{}\\
\text{}\textasciigrave{}\textasciigrave{}\textasciigrave{}yaml\\
\text{}sheet\_name: Leave used by employee\\
\text{}table\_name: Leave Entries\\
\text{}table\_range: A4:U194\\
\text{}data\_range: A5:U194\\
\text{}notes:\\
\text{~~}- "Title: Leave used summary from PP 2020-20 (Sep 27, 2020) to PP 2021-15 (July 31, 2021)"\\
\text{~~}- "Note: what is the formula to sum total for each employee under each category?"\\
\text{}header\_format: column-only\\
\text{}row\_header:\\
\text{~~}node\_1:\\
\text{~~~~}children: []\\
\text{~~~~}end\_index: A50\\
\text{~~~~}start\_index: A5\\
\text{~~~~}value: Employee 1\\
\text{~~}node\_2:\\
\text{~~~~}children: []\\
\text{~~~~}end\_index: A98\\
\text{~~~~}start\_index: A51\\
\text{~~~~}value: Employee 2\\
\text{~~}node\_3:\\
\text{~~~~}children: []\\
\text{~~~~}end\_index: A158\\
\text{~~~~}start\_index: A99\\
\text{~~~~}value: Employee 3\\
\text{~~}node\_4:\\
\text{~~~~}children: []\\
\text{~~~~}end\_index: A194\\
\text{~~~~}start\_index: A159\\
\text{~~~~}value: Employee 4\\
\text{}column\_header:\\
\text{~~}node\_1:\\
\text{~~~~}children: []\\
\text{~~~~}end\_index: A4\\
\text{~~~~}start\_index: A4\\
\text{~~~~}value: User\\
\text{~~}node\_2:\\
\text{~~~~}children: []\\
\text{~~~~}end\_index: B4\\
\text{~~~~}start\_index: B4\\
\text{~~~~}value: PP\\
\text{~~}node\_3:\\
\text{~~~~}children: []\\
\text{~~~~}end\_index: C4\\
\text{~~~~}start\_index: C4\\
\text{~~~~}value: Date\\
\text{~~}...\\
\text{}data\_properties:\\
\text{~~}User:\\
\text{~~~~}format: plain\\
\text{~~~~}type: string\\
\text{~~~~}unit: ''\\
\text{~~}PP:\\
\text{~~~~}format: plain\\
\text{~~~~}type: string\\
\text{~~~~}unit: ''\\
\text{~~}Date:\\
\text{~~~~}format: date-string\\
\text{~~~~}type: date\\
\text{~~~~}unit: ''\\
\text{~~}...\\
\text{}\textasciigrave{}\textasciigrave{}\textasciigrave{}\\
\text{}\\
\text{}\#\#\# Designated Summary Area\\
\text{}\\
\text{}\textasciigrave{}\textasciigrave{}\textasciigrave{}yaml\\
\text{}sheet\_name: Leave used by employee\\
\text{}table\_name: Leave Summary Area\\
\text{}table\_range: L4:U8\\
\text{}data\_range: M5:U8\\
\text{}notes:\\
\text{~~}- "This area is part of the main table but designated for summary calculations"\\
\text{~~}- "Intended for formula-based calculations by employee and leave type"\\
\text{~~}- "Currently contains only row headers (L5:L8) and column headers (M4:U4)"\\
\text{}header\_format: both-row-and-column\\
\text{}row\_header:\\
\text{~~}node\_1:\\
\text{~~~~}children: []\\
\text{~~~~}end\_index: L5\\
\text{~~~~}start\_index: L5\\
\text{~~~~}value: Employee 1\\
\text{~~}node\_2:\\
\text{~~~~}children: []\\
\text{~~~~}end\_index: L6\\
\text{~~~~}start\_index: L6\\
\text{~~~~}value: Employee 2\\
\text{~~}node\_3:\\
\text{~~~~}children: []\\
\text{~~~~}end\_index: L7\\
\text{~~~~}start\_index: L7\\
\text{~~~~}value: Employee 3\\
\text{~~}node\_4:\\
\text{~~~~}children: []\\
\text{~~~~}end\_index: L8\\
\text{~~~~}start\_index: L8\\
\text{~~~~}value: Employee 4\\
\text{}column\_header:\\
\text{~~}node\_1:\\
\text{~~~~}children: []\\
\text{~~~~}end\_index: M4\\
\text{~~~~}start\_index: M4\\
\text{~~~~}value: Annual Leave\\
\text{~~}node\_2:\\
\text{~~~~}children: []\\
\text{~~~~}end\_index: N4\\
\text{~~~~}start\_index: N4\\
\text{~~~~}value: LWOP\\
\text{~~}node\_3:\\
\text{~~~~}children: []\\
\text{~~~~}end\_index: O4\\
\text{~~~~}start\_index: O4\\
\text{~~~~}value: Sick Leave\\
\text{~~}...\\
\text{}data\_properties:\\
\text{~~}Annual Leave:\\
\text{~~~~}format: decimal\\
\text{~~~~}type: empty\\
\text{~~~~}unit: hours\\
\text{~~}LWOP:\\
\text{~~~~}format: decimal\\
\text{~~~~}type: empty\\
\text{~~~~}unit: hours\\
\text{~~}Sick Leave:\\
\text{~~~~}format: decimal\\
\text{~~~~}type: empty\\
\text{~~~~}unit: hours\\
\text{~~}...\\
\text{}\textasciigrave{}\textasciigrave{}\textasciigrave{}\\
\end{tcolorbox}

\begin{tcolorbox}[
    breakable,
    float*=h,
    width=\textwidth, 
    colback=background, 
    colframe=title, 
    fonttitle=\bfseries, 
    title=Verification Example
]
\text{}verification: false\\
\text{}issues:\\
\text{~~~~}- "The 'Leave Entries' table range is defined as A4:U194, but the actual data ends before row 194, leaving many blank rows at the end."\\
\text{~~~~}- "In the 'Leave Summary Area', the range is defined as L4:U8, but column L contains employee names instead of being part of the summary calculation area. This misaligns the row headers for the summary section."\\
\text{~~~~}- "Column headers for the 'Leave Entries' table are correctly present in row 4, but the 'Leave Summary Area' overlaps with the main table without clear structural separation."\\
\text{~~~~}- "Some cells in the 'Leave Entries' table contain unexpected values (e.g., employee names in column M), suggesting misplaced data or incorrect usage of summary columns (M-U)."\\
\text{~~~~}- "The note 'what is the formula to sum total for each employee under each category?' appears in row 3, indicating incomplete setup for summary calculations in the designated area." \\
\end{tcolorbox}

\end{document}